\def\year{2018}\relax
\documentclass[letterpaper]{article} 
\usepackage{aaai18}  
\usepackage{times}  
\usepackage{helvet}  
\usepackage{courier}  
\usepackage{url}  

\usepackage{graphicx}  
\usepackage{amsmath}
\usepackage{amsfonts}       
\usepackage{color}
\usepackage{booktabs}       
\usepackage[sort,nocompress]{cite}

\begin{document}

\title{Towards Ultra-High Performance and Energy Efficiency of Deep Learning Systems: An Algorithm-Hardware Co-Optimization Framework}

\author{
Yanzhi Wang\textsuperscript{1}, Caiwen Ding\textsuperscript{1}, Zhe Li\textsuperscript{1}, Geng Yuan\textsuperscript{1}, Siyu Liao\textsuperscript{2},\\
{\bf \Large Xiaolong Ma\textsuperscript{1}, Bo Yuan\textsuperscript{2}, Xuehai Qian\textsuperscript{3}, Jian Tang\textsuperscript{1}, Qinru Qiu\textsuperscript{1}, Xue Lin\textsuperscript{4}}\\
\textsuperscript{1}Department of Electrical Engineering and Computer Science, Syracuse University, Syracuse, NY 13244\\
\textsuperscript{2}Department of Electrical Engineering, City University of New York, New York, NY 10031\\
\textsuperscript{3}Department of Electrical Engineering, University of Southern California, Los Angeles, CA 90089\\
\textsuperscript{4}Department of Electrical and Computer Engineering, Northeastern University, Boston, MA 02115\\
}



\maketitle

\begin{abstract}
Hardware accelerations of deep learning systems have been extensively investigated in industry and academia. The aim of this paper is to achieve ultra-high energy efficiency and performance for hardware implementations of deep neural networks (DNNs). An algorithm-hardware co-optimization framework is developed, which is applicable to different DNN types, sizes, and application scenarios. The algorithm part adopts the general block-circulant matrices to achieve a fine-grained tradeoff of accuracy and compression ratio. It applies to both fully-connected and convolutional layers and contains a mathematically rigorous proof of the effectiveness of the method. The proposed algorithm reduces computational complexity per layer from O($n^2$) to O($n\log n$) and storage complexity from O($n^2$) to O($n$), both for training and inference. The hardware part consists of highly efficient \emph{Field Programmable Gate Array} (FPGA)-based implementations using effective reconfiguration, batch processing, deep pipelining, resource re-using, and hierarchical control. Experimental results demonstrate that the proposed framework achieves at least 152X speedup and 71X energy efficiency gain compared with IBM TrueNorth processor under the same test accuracy. It achieves at least 31X energy efficiency gain compared with the reference FPGA-based work.
\end{abstract}

\section{Introduction}
The recent \emph{deep neural networks} (DNNs), especially deep \emph{convolutional neural networks} (CNNs), have been able to deliver remarkable success in visual and recognition tasks (\citeauthor{deng2009imagenet,taigman2014deepface}) and real-world applications (\cite{huval2015empirical,collobert2008unified,burbidge2001drug}), by leveraging large-scale neural network sizes and learning from a huge volume of data. Despite the advantage of improved overall accuracy, the deep layered structure and large model sizes increase the computational complexity and memory requirements. 
It is projected that the majority of inference tasks will be performed on embedded, IoT and mobile systems which are with limited power and computational resources. 
In order to achieve higher scalability, performance, and energy efficiency, two orthogonal research and development trends have both attracted enormous interests.

The first is hardware accelerations of deep learning systems/applications, which have been extensively investigated in industry and academia (\cite{farabet2009cnp,suda2016throughput,qiu2016going,zhang2016caffeine,zhang2016energy,han2017ese,zhao2017accelerating,zhang2017improving,umuroglu2016finn,company1,company2,chen2017eyeriss,han2016eie,chen2014diannao}). As a representative technique, \underline{FPGA-based accelerators} can offer the advantages of programmability, high degree of parallelism and short development cycle. Important progresses have been reported on FPGA accelerations of original DNNs (\cite{farabet2009cnp,suda2016throughput,zhang2016caffeine,zhang2016energy}), binary neural networks (\cite{zhao2017accelerating,umuroglu2016finn}), and more recently, on DNNs and recurrent neural networks (RNNs) with model compression techniques (\cite{qiu2016going,han2017ese}). These prior work mainly focus on the inference phase of DNNs, and suffer from frequent access to off-chip memory systems because the limited on-chip memory can hardly accommodate the large model sizes. Accessing off-chip memory is highly energy inefficient. As pointed out in (\cite{han2015learning,han2015deep}), the per-bit access energy of off-chip memory is 200X compared with on-chip memory storage, and dominates the whole system power consumptions. Besides, it is also desirable to achieve algorithmic-level accelerations to accommodate the further scaling of DNNs, instead of simply adding more and more hardware devices.

The second important trend is the model size compression and algorithmic-level acceleration of DNNs (with very minor accuracy loss), including weight quantization (\cite{lin2016fixed,lin2015neural}), sparsity regularization (\cite{feng2015learning,wen2016learning,li2017enabling}), connection pruning (\cite{han2015learning,han2015deep}), and low rank approximation (\cite{denil2013predicting,denton2014exploiting}). These approaches can offer a reasonable amount of parameter reduction (e.g., by $9\times$ to $13\times$ in (\cite{han2015learning,han2015deep})) and/or a reasonable speedup (e.g., around 50\% to 2$\times$ in (\cite{wen2016learning})). However, they suffer from the following limitations: (i) the sparsity regularization and pruning methods will likely result in an irregular and sparse network structure, thereby undermining the compression ratio and increasing computation time (especially inefficient on GPUs and dedicated hardware which has high parallelism capability); (ii) the training complexity will be increased by incorporating additional pruning process (\cite{han2015learning,han2015deep}), additional low rank approximation step (\cite{denil2013predicting,denton2014exploiting}), or extra trade-off parameters (\cite{wen2016learning}); (iii) the compression or acceleration factors are heuristic numbers that cannot be precisely controlled, not to mention a mathematically rigorous proof of the effectiveness of these methods.

To combine these two directions, the aim of this paper is to address the limitations of existing model size compression and acceleration work and to achieve ultra-high energy efficiency and performance for FPGA-based hardware implementations of DNNs, by (i) deriving a highly suitable algorithm for efficient computation and storage reduction without significant accuracy loss, and (ii) deriving the corresponding optimized hardware implementations.
We develop an algorithm-hardware co-optimization framework, which is applicable to different DNN types, sizes, and application scenarios. The proposed framework comprises algorithm and hardware parts. The \textbf{algorithm part} extends reference (\cite{cheng2015exploration}), which applies circulant matrices to the whole fully-connected (FC) layer for model compression, to (i) the adoption of the general \emph{block-circulant matrices} to achieve fine-grained tradeoff of accuracy and compression ratio, (ii) the generalization to the convolutional (CONV) layers for significant acceleration as CONV layers dominate the computation of DNNs (\cite{krizhevsky2012imagenet,he2016deep}), (iii) providing a mathematically rigorous proof that the proposed algorithm will asymptotically converge to the same ``effectiveness'' as DNNs without compression, and (iv) decoupling the fast Fourier transform (FFT) and inverse FFT computations in the framework for accelerating computation and facilitating hardware implementations. The proposed algorithm reduces computational complexity per layer from O($n^2$) to O($n\log n$) and storage complexity from O($n^2$) to O($n$), both for training and inference, with negligible degradation in DNN accuracy. The \textbf{hardware part} consists of highly efficient FPGA-based implementations using effective reconfiguration, batch processing, deep pipelining technique, effective resource re-using, and a hierarchical control framework. The proposed FPGA-based implementation can accommodate the whole DNN model using on-chip block memory, thereby significantly improving the overall energy efficiency. Finally, a comprehensive \textbf{algorithm-hardware co-optimization} is proposed which comprises (i) model selection and optimization, (ii) hardware optimization, and (iii) variational inference-based Bayesian learning for enhancing accuracy and robustness.
In summary, the major contributions of this work include both algorithm and hardware parts. The algorithm part adopts block-circulant matrices for weight representation, which could achieve a significant model compression ratio with minor accuracy degradation. It applies to the whole network, both fully-connected and convolutional layers. The hardware part consists of highly efficient FPGA-based implementations with multiple innovative parts of reconfiguration, batch processing, deep pipelining, resource re-using, etc. 

Please note that the proposed framework \textbf{is distinct from} the prior work (\cite{mathieu2013fast}), which applies FFTs to accelerate the computations in the CONV layers. The prior work applies only to a single filter in the CONV layer and achieves no storage reduction (in fact it results in storage increase), whereas the proposed method applies both to CONV and FC layers and achieves simultaneous acceleration and storage reduction.

Because we focus on highly energy-efficient FPGA-based implementations for low-power embedded applications, we focus on the inference phase of small to medium-scale DNNs (e.g., for MNIST, SVHN, CIFAR datasets) on high energy-efficiency FPGAs. Compared with the IBM TrueNorth neurosynapstic processor (\cite{merolla2014million}), our FPGA-based implementation achieves at least 152X speedup in throughput and 71X energy efficiency gain under the same test accuracy. Similarly, our actual FPGA implementations outperform (in performance) the state-of-the-art analog-based and emerging device-based implementations. Our framework achieves at least 31X gain in equivalent energy efficiency compared with the reference FPGA-based work that achieves the best efficiency.

\section{Related Works}

\emph{\textbf{FPGA Accelerations of DNNs.}} FPGA-based accelerations of DNNs have been extensively investigated recently due to the advantages of programmability, high degree of parallelism and short development cycle. Based on the early work of direct acceleration of FPGAs (\cite{farabet2009cnp}), recently researchers have investigated energy-efficient implementations using the batch processing technique (\cite{zhang2016caffeine,zhang2016energy}) or on compressed models using singular value decomposition (SVD) (\cite{qiu2016going}). In this year the research on this topic has exploded, including accelerations of DNNs with weight pruning (\cite{han2017ese}), binary neural networks (\cite{zhao2017accelerating,umuroglu2016finn}), and high-level synthesis for fast generation of FPGA implementations (\cite{zhao2017accelerating,zhang2017improving}). These work typically suffer from frequent access to off-chip memory systems because their model sizes cannot be effectively reduced for on-chip memory storage, thereby resulting in high energy consumptions. The typical (equivalent) energy efficiency range is from 7 GOPS/W to less than 1 TOPS/W, depending on the testing FPGA platform, implementation details, and compression techniques.

\emph{\textbf{Connection Pruning and Weight Sparsifying.}} Han \emph{et al.} (\cite{han2015learning,han2015deep}) reduced the number of parameters by 9X - 13X using connection pruning. Since most reduction is achieved on FC layers, no significant speedups of CONV layers can be observed (\cite{wen2016learning}). As CONV layers have become the computational bottleneck, compression and acceleration on CONV layers become essential. Liu \emph{et al.} achieved layer-wise 4.59X speedup on the CONV layers of \emph{AlexNet} with 2\% accuracy loss. Recently, (\cite{wen2016learning}) adopts a \emph{structured sparsity learning} method and derives an effective tradeoff between acceleration on CPU/GPU and test accuracy for the CONV layers. More specifically, for \emph{ResNet-20} on CIFAR-10 and \emph{AlexNet} on ImageNet benchmarks, more than 50\% acceleration can be achieved without any accuracy loss, while around 3X acceleration is achieved with an acceptable accuracy loss of 2\%.

\emph{\textbf{FFTs for CONV Layer Accelerations.}} LeCun \emph{et al.} have proposed using FFTs to accelerate the computations in the CONV layers, which applies only to a single filter in the CONV layer (\cite{mathieu2013fast}). It uses FFT to calculate the traditional inner products of filters and input feature maps, and can achieve speedup for large filter sizes. The underlying neural network structure remains \emph{unchanged}. The speedup is due to filter reuse and it cannot achieve either asymptotic speedup in Big-O notation or weight compression.

\emph{\textbf{Structured Matrices in FC Layers for Model Compression.}} The most relevant work to this paper is (\cite{cheng2015exploration}), which directly applies circulant matrices to the FC layers for model compression. As an example, an FC layer of DNN can be represented as $\mathbf{y}=\psi(\mathbf{Wx}+\mathbf{\theta})$, where vectors $\mathbf{x}$ and $\mathbf{y}$ represent the outputs of all neurons in the previous layer and the current layer, respectively; $\mathbf{W}$ is an $n$-by-$n$ weight matrix; and $\psi(\cdot)$ is the activation function. When $\mathbf{W}$ is a circulant matrix, the fast Fourier transform (FFT)-based fast multiplication method can be utilized, and the computational complexity and weight storage complexity will be reduced from O($n^2$) to O($n\log n$) and from O($n^2$) to O($n$), respectively. Despite the significant reduction in computation and weight storage, this approach has the limitations of (i) resulting in a huge number of padding 0's when the numbers of inputs and outputs are not equal, (ii) resulting in certain accuracy degradation for large-scale FC layers because of the aggressive weight reduction, and (iii) only applicable to the FC layer, whereas the CONV layers are the most computationally intensive in DNNs.

\section{Algorithm Development of Block-Circulant Matrix-Based DNNs}

In this section, we develop the algorithmic framework of block-circulant matrix-based DNNs for simultaneous acceleration and model compression, for both inference and training phases. The proposed framework is able to accommodate arbitrary size and aspect ratio of weight matrices, and achieves a fine-grained tradeoff between test accuracy and compression/acceleration ratio (\cite{ding2017c}). Unlike (\cite{cheng2015exploration}), we develop algorithms for both FC and CONV layers as shown in the following. We provide a mathematically rigorous proof of the proposed algorithm that it satisfies the \emph{universal approximation property} as uncompressed DNNs. Finally, we develop decoupling technique for FFT/IFFT pairs for further acceleration and facilitating hardware (FPGA) implementations.

\subsection{Inference and Training Algorithms for FC Layers}

\begin{figure}
  \centering
  \includegraphics[width=0.3\textwidth]{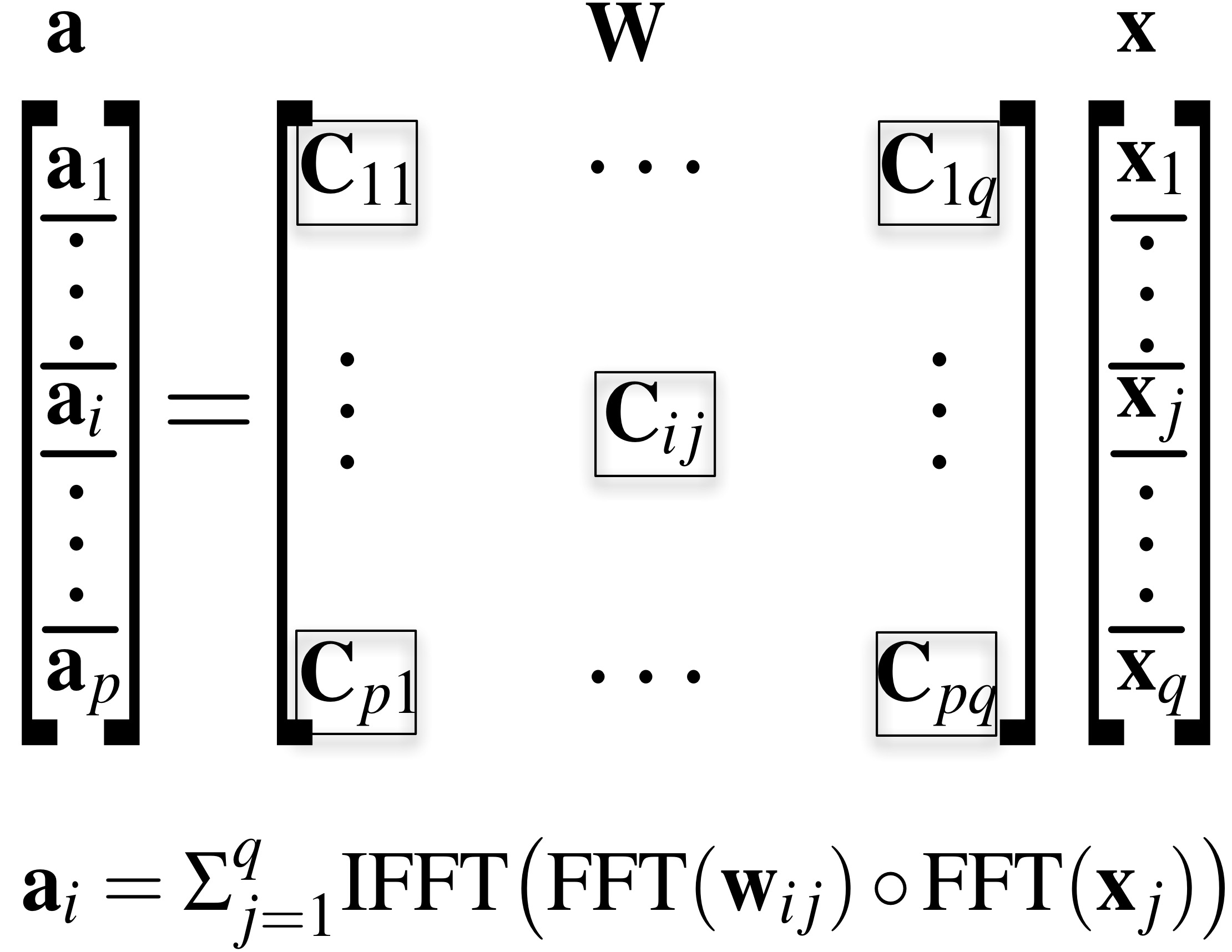}
  \caption{\footnotesize\bf Block-circulant matrix-vector multiplication in the proposed framework.}
  \label{fig:block}
\end{figure}

The key idea of block-circulant matrix-based FC layers is to partition the original arbitrary-size unstructured weight matrix $\textbf{W}\in\mathbb{R}^{m\times n}$ into 2D blocks of square sub-matrices. Such partitioning strategy has two advantages: 1) It is suitable for arbitrary-size weight matrices without any requirement on the aspect ratio of $\textbf{W}$; and 2) it is an adjustable approach that can conveniently control the compression ratio and potential accuracy loss by only changing the size of sub-matrices.

For formal discussions on the proposed inference and training procedures, let $k$ denote the \emph{block size} (size of each sub-matrix) and there are $p \times q$ blocks after partitioning $\mathbf{W}$, where $p = m\div k$ and $q=n \div k$. Zero padding is required if $k$ does not directly divide $m$ or $n$, but the amount of zero padding will be significantly reduced compared with (\cite{cheng2015exploration}). Then $\mathbf{W} = [\mathbf{C}_{ij}]$, $i \in \{1 \dots p\}$, $j \in \{1 \dots q\}$. Correspondingly, the input $\mathbf{x}$ is also partitioned as $\mathbf{x} = [\mathbf{x}^T_1, \mathbf{x}^T_2, \dots, \mathbf{x}^T_q]^T$. Then the forward propagation process in the inference phase is given by:
\begin{equation}
\mathbf{a}
=
\mathbf{Wx} 
=
\begin{bmatrix}
         \sum_{j=1}^q \mathbf{C}_{1j} \mathbf{x}_j   \\
         \sum_{j=1}^q \mathbf{C}_{2j} \mathbf{x}_j   \\
         \dots \\
         \sum_{j=1}^q \mathbf{C}_{pj} \mathbf{x}_j  
\end{bmatrix}
=
\begin{bmatrix}
         \mathbf{a}_1   \\
         \mathbf{a}_2   \\
         \dots \\
         \mathbf{a}_p
\end{bmatrix},
\end{equation}
where $\mathbf{a}_i \in \mathbb{R}^{k}$ is a column vector. Assume each circulant matrix $\mathbf{C}_{ij}$ is defined by a vector $\mathbf{w}_{ij}$, i.e., $\mathbf{w}_{ij}$ is the first row vector of $\mathbf{C}_{ij}$. Then according to the \emph{circulant convolution theorem} (\cite{pan2012structured,bini1996polynomial}), the calculation of $\mathbf{C}_{ij} \mathbf{x}_j$ can be performed as $\text{IFFT}\big(\text{FFT}(\mathbf{w}_{ij})\circ\text{FFT}(\mathbf{x}_j)\big)$, where $\circ$ denotes element-wise multiplications. The operation procedure is shown in Fig. \ref{fig:block}. For the inference phase, the computational complexity of this FC layer will be $O(pqk\log k)$, which is equivalent to $O(n\log n)$ for small $p$, $q$ values. Similarly, the storage complexity will be $O(pqk)$ because we only need to store $\mathbf{w}_{ij}$ or $\text{FFT}(\mathbf{w}_{ij})$ for each submatrix, which is equivalent to $O(n)$ for small $p$, $q$ values. Simultaneous acceleration and model compression compared with the original DNN can be achieved.

Now consider the backward propagation process in the training phase. Let $a_{il}$ be the $l$-th output element in $\mathbf{a}_i$. Then by using the chain  rule we can derive the backward propagation process as follows:
\begin{equation}
\frac{\partial L}{\partial \mathbf{w}_{ij}} 
= \sum_{l=1}^k \frac{\partial L}{\partial a_{il}} \frac{\partial a_{il}}{\partial \mathbf{w}_{ij}}
= \frac{\partial L}{\partial \mathbf{a}_i} \frac{\partial \mathbf{a}_i}{\partial \mathbf{w}_{ij}},
\end{equation}
\begin{equation}
\frac{\partial L}{\partial \mathbf{x}_{j}} 
= \sum_{i=1}^p\sum_{l=1}^k \frac{\partial L}{\partial a_{il}} \frac{\partial a_{il}}{\partial \mathbf{x}_{j}}
= \sum_{i=1}^p \frac{\partial L}{\partial \mathbf{a}_i} \frac{\partial \mathbf{a}_i}{\partial \mathbf{x}_j}.
\end{equation}
We have proved that $\frac{\partial \mathbf{a}_i}{\partial \mathbf{w}_{ij}}$ and $\frac{\partial\mathbf{a}_i}{\partial \mathbf{x}_j}$ are block-circulant matrices. Therefore, $\frac{\partial L}{\partial \mathbf{w}_{ij}}$ and $\frac{\partial L}{\partial \mathbf{a}_i} \frac{\partial \mathbf{a}_i}{\partial \mathbf{x}_j}$ can be calculated as the ``FFT$\rightarrow$element-wise multiplication$\rightarrow$IFFT'' procedure and is equivalent to $O(n\log n)$ computational complexity per layer. Due to space limitation, the algorithmic descriptions of forward and backward propagations are omitted.

Please note that there is no special need to translate into or approximate each sub-matrix of $\mathbf{W}$. Instead, as shown in Eqns. (2) and (3), we directly learn the vector $\mathbf{w}_{ij}$ (the first-row vector) of each sub-matrix of $\mathbf{W}$ in the training process. The assumption is that the other rows of the sub-matrix follow the circulant formulation. In other words, when following the learning process Eqns. (2) and (3), the learnt weight matrices naturally follow the block-circulant format. In fact, this is a key advantage of this proposed method in that there is no need for additional ``translation" or ``approximation" steps. 

\subsection{Inference and Training for CONV Layers}
We generalize the inference and training algorithms to CONV layers, which have become the computation bottleneck of the whole DNN. The CONV layers are often associated with multiple input and multiple output feature maps:
\begin{equation}\label{eqn:tensor}
\mathcal{Y}(x,y,p)=\sum_{i=1}^r\sum_{j=1}^r\sum_{c=1}^C\mathcal{F}(i,j,c,p)\mathcal{X}(x+i-1,y+j-1,c),
\end{equation}
where $\mathcal{X}\in\mathbb{R}^{W\times H\times C}$, $\mathcal{Y}\in\mathbb{R}^{(W-r+1)\times(H-r+1)\times P}$, $\mathcal{F}\in\mathbb{R}^{r\times r\times C\times P}$ represent the input, output, and weight tensors of the CONV layer, respectively. Here $W$ and $H$ are the spatial dimensions of the input feature maps, $C$ is the number of input feature maps, $r$ is the size of the convolutional kernel, and $P$ is the number of output feature maps.

\begin{figure}
  \centering
  \includegraphics[width=0.35\textwidth]{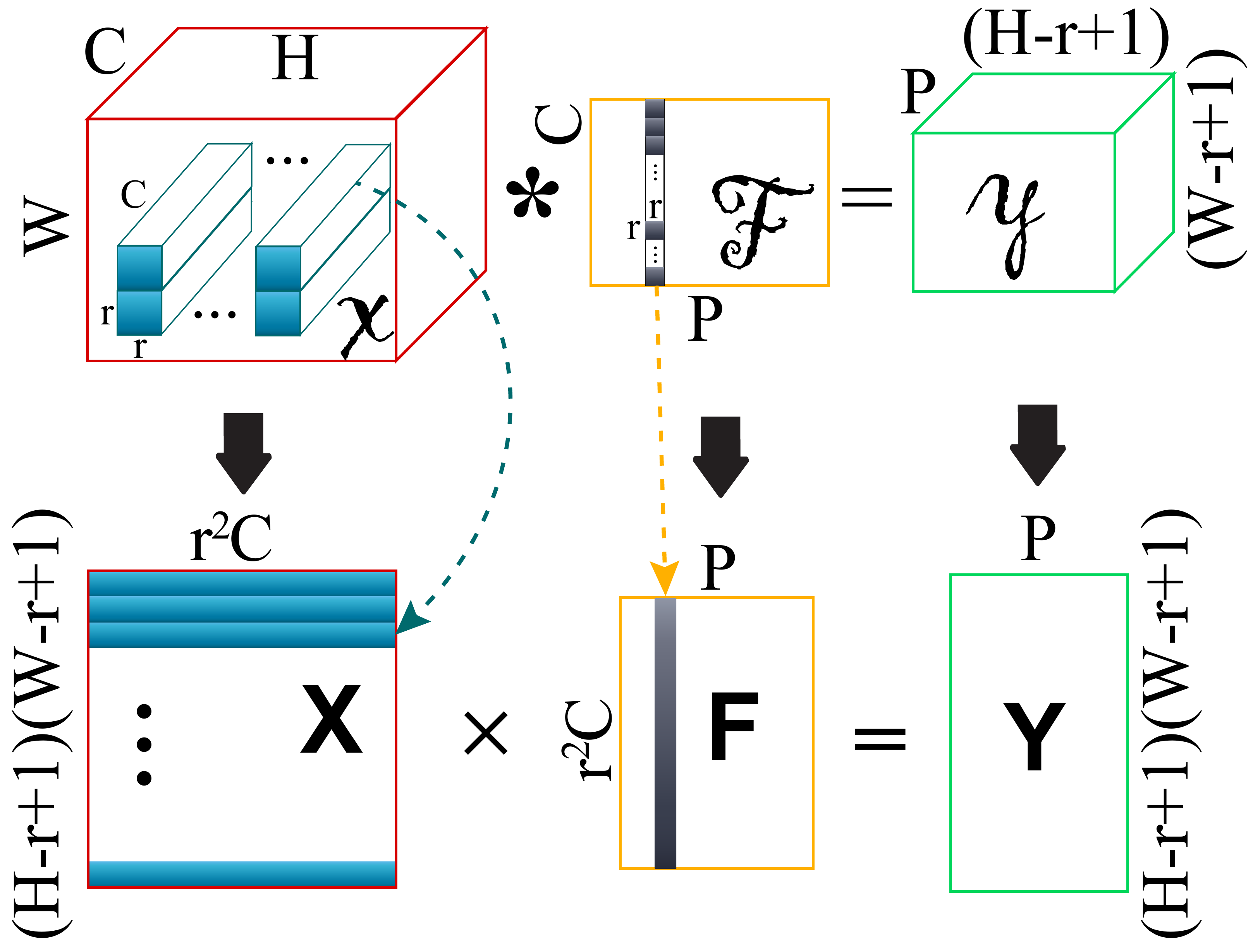}
  \caption{\footnotesize\bf Reformulation of Eqn. (\ref{eqn:tensor}) to matrix multiplication.}
  \label{fig:reformulation}
\end{figure}

Efficient software tools such as Caffe provide an efficient methodology of transforming tensor-based operations in the CONV layer to matrix-based operations (\cite{jia2014caffe,vedaldi2015matconvnet}), in order to enhance the implementation efficiency (GPUs are optimized for matrix operations.) Fig. \ref{fig:reformulation} illustrates the application of the method to reformulate Eqn. (4) to the matrix multiplication $\mathbf{Y}=\mathbf{XF}$, where $\mathbf{X}\in\mathbb{R}^{(W-r+1)(H-r+1)\times Cr^2}$, $\mathbf{Y}\in\mathbb{R}^{(W-r+1)(H-r+1)\times P}$, and $\mathbf{F}\in\mathbb{R}^{Cr^2\times P}$.

We generalize the concept of ``block-circulant structure" to the rank-4 tensor ($\mathcal{F}$) in the CONV layer, i.e., \emph{all the slices of the form $\mathcal{F}(\cdot,\cdot,c,p)$ are block-circulant matrices.} Then we can prove that $\mathbf{F}$ is actually a block-circulant matrix. 
Hence the fast multiplication approach for block-circulant matrices, as the ``FFT$\rightarrow$component-wise multiplication $\rightarrow$IFFT" procedure, can now be applied to accelerate $\mathbf{Y}=\mathbf{XF}$, thereby resulting in the acceleration of (4). The training phase can be derived similarly. The overall degrees of reduction in computational and storage complexities are similar to those in FC layers. 

\subsection{Theoretical Foundation and Software Results}
With the substantial reduction of weight storage and computations, we also attempt to prove that the proposed block-circulant matrix-based framework will consistently yield the similar overall accuracy compared with DNNs without compression. The theoretical proof will make the proposed method theoretically rigorous and distinct from prior work.

In the theory of neural networks, the \emph{universal approximation property} states that a neural network should be able to approximate any continuous or measurable function with arbitrary accuracy provided that an enough large number of parameters are available. This property provides the theoretical guarantee of using neural networks to solve machine learning problems, since machine learning tasks can be formulated as finding a proper approximation of an unknown, high-dimensional function. We have proved the \underline{universal approximation property of bl-} \underline{ock circulant matrix-based neural networks}, and more generally, for arbitrary structured matrices satisfying the low displacement rank $\gamma$. As a result, we can guarantee the universal ``effectiveness" of the proposed framework on different DNN types and sizes, application domains, and hardware/software platforms. Detailed proof procedure is provided in the supplementary file (\cite{proof_simple}).

\begin{figure}
  \centering
  \includegraphics[width=0.45\textwidth]{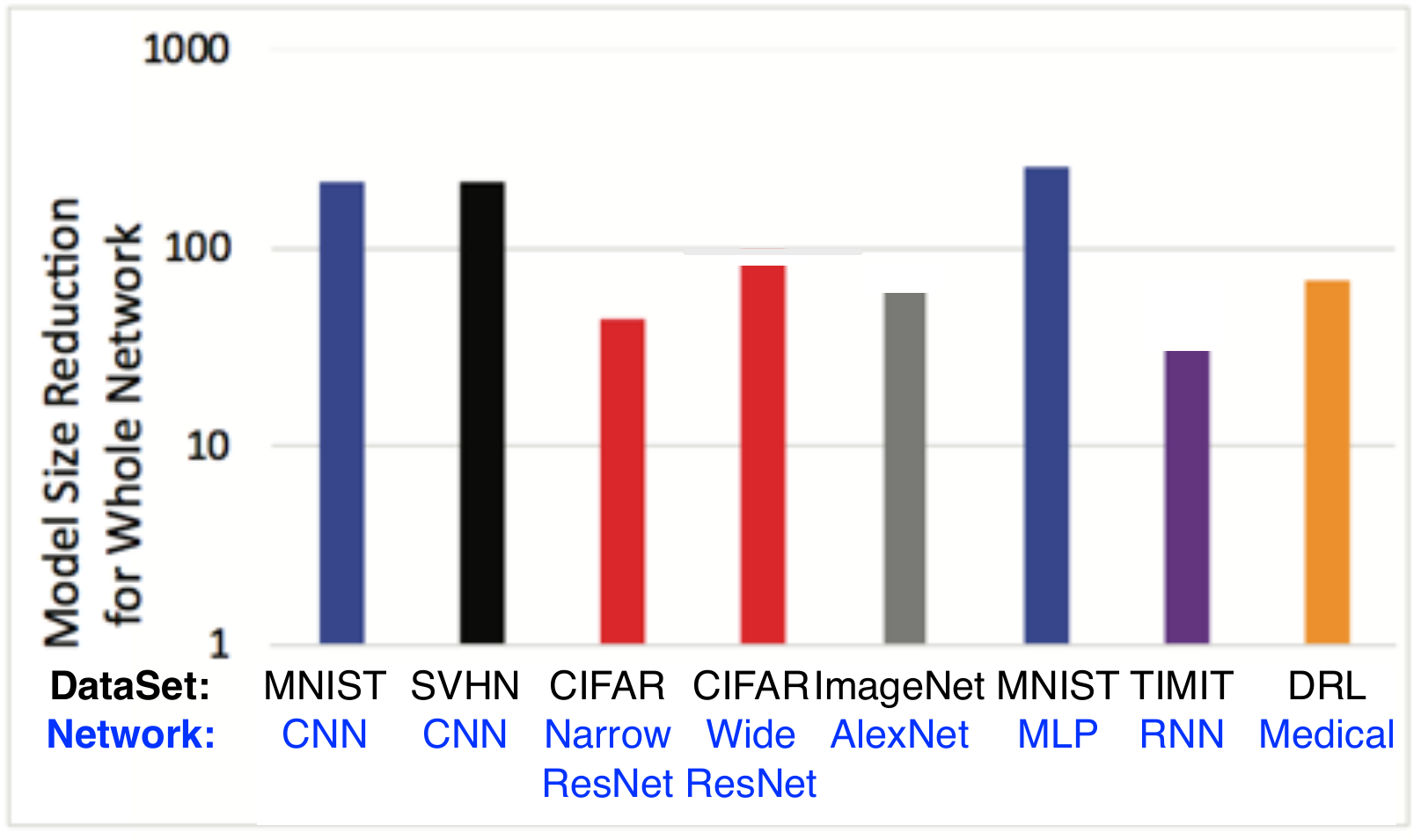}
  \caption{\footnotesize\bf Weight storage reduction results.}
  \label{fig:compress}
\end{figure}

Fig. \ref{fig:compress} shows the model compression results on MNIST, SHVN, CIFAR-10, ImageNet, TIMIT (speech recognition) benchmarks, etc., using various DNN models. The accuracy degradations are constrained to be 1\% to 2\% between the original models and block-circulant matrix-based models. The overall model compression is contributed by both weight parameter reduction and bit quantization. It can be observed that a significant model size compression, and therefore acceleration, can be achieved using the proposed framework. 

\subsection{Accelerating Computation and Facilitating Hardware Implementations}

We propose the decoupling technique of FFTs and IFFTs, which applies to both inference and training phases. We take the inference phase of FC layer as an illustrative example. First, we make the observation that the FFT results of $\mathbf{x}_j$, i.e., $\text{FFT}(\mathbf{x}_j)$, need to be utilized to calculate all $\mathbf{a}_i$ vectors. Similar observation also holds for $\mathbf{w}_{ij}$. Hence, we could perform pre-calculation of $\text{FFT}(\mathbf{x}_j)$ and $\text{FFT}(\mathbf{w}_{ij})$ and store them in memory for effective re-use. The $\text{FFT}(\mathbf{w}_{ij})$ values can even be pre-calculated and stored in memory before the inference phase because they are fixed after training. By performing such pre-calculation of $\text{FFT}(\mathbf{x}_j)$, the total number of FFTs needed to calculate $\mathbf{Wx}$ reduces from $p\cdot q$ to $q$ (assuming $\text{FFT}(\mathbf{w}_{ij})$'s are calculated and stored in prior), achieving a significant reduction in total computations.

Similarly, each vector $\mathbf{a}_i$ to be calculated in Eqn. (1) is given by $\sum_{j=1}^q \text{IFFT}\big(\text{FFT}(\mathbf{w}_{ij})\circ\text{FFT}(\mathbf{x}_j)\big)$, which requires $q$ IFFT calculations. Because FFTs and IFFTs are linear operations (\cite{oppenheim1999discrete}), we can calculate IFFT in the last step, i.e., calculate $\mathbf{a}_i$ as $\text{IFFT}\big(\sum_{j=1}^q \text{FFT}(\mathbf{w}_{ij})\circ\text{FFT}(\mathbf{x}_j)\big)$. In this way the total number of IFFT calculations can be reduced by $q$ times.

\section{High Energy Efficiency and Performance Implementation in FPGAs}

Based on the algorithmic framework, we describe the developed high-efficiency FPGA-based implementation of DNNs. Since the target is 
low-power embedded applications, we focus on the inference phase of small to medium-scale DNNs, e.g., for MNIST, SVHN, CIFAR datasets. We leave the large-scale DNNs, e.g., for ImageNet dataset, for future investigation because they do not target at embedded applications. We first describe the proposed FPGA implementations using a set of reconfiguration and performance/efficiency enhancement techniques, then present the algorithm-hardware co-optimization framework. 

\subsection{FPGA Implementations: Reconfigurability, In-Place Computation, Batch Processing, Deep Pipelining, and Resource Re-Use}

\emph{\textbf{Reconfigurability, In-Place Computation, and Batch Processing.}} In order to accommodate different DNN models, sizes, and application scenarios, the proposed FPGA implementation possesses \emph{reconfigurability} for different layer sizes and layer types (FC or CONV layers). The reconfigurability is achieved because (i) both FC and CONV layers are formulated as the ``FFT$\rightarrow$component-wise multiplication $\rightarrow$IFFT" procedure; (ii) IFFT can be implemented using the FFT structure with simple pre-processing step (\cite{salehi2013pipelined}); and (iii) the FFT structure possesses inherent \emph{recursive property} in that small-scale FFTs can be implemented in parallel in larger-scale FFT structures (\cite{oppenheim1999discrete}). More specifically, the first and second properties enable the implementation of a single FFT structure in a time-multiplexed manner for both FFTs and IFFTs and both FC and CONV layers. For instance, a 128-input FFT structure can be implemented in FPGA if a block size of 128 is utilized. The third property enables that a single FFT structure can be utilized even if we use different block sizes for FC and CONV layers. Finally, \emph{in-place computation} is utilized such that the same memory space can be utilized to store the outputs of every layer in the DNN, i.e., the outputs of each neuron layer $i$ will replace the inputs (outputs of layer $i-1$). In this way, the execution of an overall DNN will use the single FFT structure in a sequential, time-multiplexed manner without extra memory requirements.

\begin{figure}
  \centering
  \includegraphics[width=0.45\textwidth]{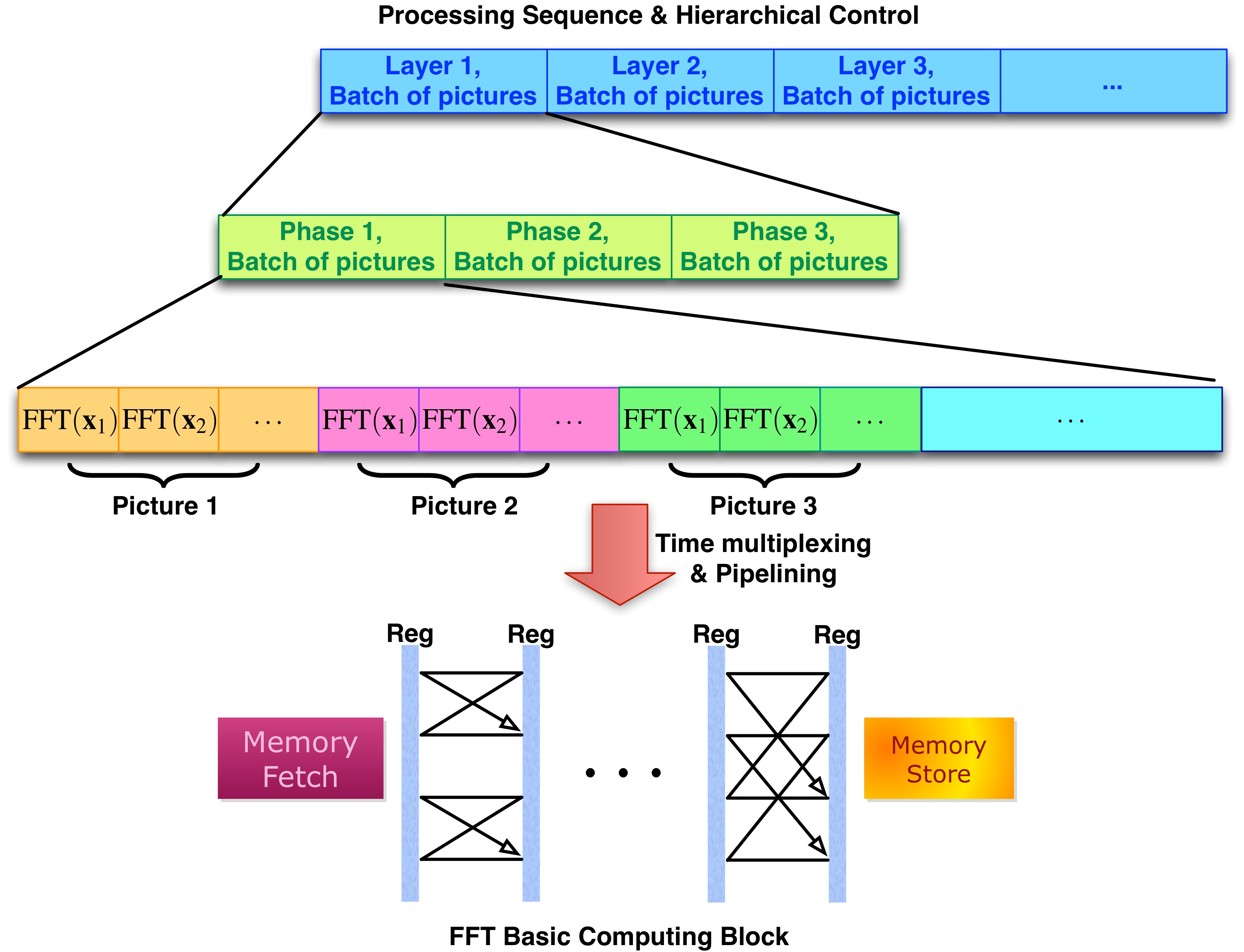}
  \caption{\footnotesize\bf The execution of the whole DNN inference phase in FPGA.}
  \label{fig:struction}
\end{figure}

The execution of the inference phase of the whole DNNs is shown in Fig. \ref{fig:struction}. The \emph{batch processing} technique is utilized, in that a batch of input pictures are processed in an interleaved manner in the FPGA. As shown in Fig. \ref{fig:struction}, we first compute the first layer of all input pictures in this batch, then the second layer, and so on. 
Different layers of a neural network will be time-multiplexed on the basic block.
The computations are all based on the implemented FFT structure discussed previously in a time-multiplexed manner. 
All operations will be pipelined on the basic computing block.
The reason of batch processing is the \emph{deep pipelining} (to be discussed later) utilized in the hardware implementation. Otherwise, pipeline bubbles have to be injected when computing all layers for one input picture consecutively, which results in timing overheads. A typical batch consists of around 50-100 pictures, because (i) state-of-the-art FPGAs have more than 2MB on-chip memory storage (e.g., Intel (Altera) CyClone V 5CEA9, Xilinx Kintex-7 XC7K325T) and (ii) the intermediate results of small to medium-scale DNNs (e.g., DNNs for CIFAR-10) typically take several KBs per picture.

\emph{\textbf{Three-Phase Operations, Deep Pipelining, and Resource Re-Use.}} As described before, the calculation of $\mathbf{Wx}$ consists of three phases: calculation of $\text{FFT}(\mathbf{x}_j)$ vectors for each $j$, calculation of element-wise multiplications $\text{FFT}(\mathbf{w}_{ij})\circ\text{FFT}(\mathbf{x}_j)$ for each $i$, $j$ (and corresponding additions), and IFFTs for each $i$. For example, if $\mathbf{W}$ is 1024-by-1024 and the block size is 128, a total of 8 FFTs, 8 IFFTs, and 64 groups of element-wise multiplications will be performed. As shown in Fig. \ref{fig:struction}, the three-phase operations are integrated with batch processing. More specifically, an outer loop iterates on all layers of the DNN. Within the outer loop is the three calculation phases. Within each phase is the calculations for every $i$, $j$ in each picture and for all pictures. In this way the timing overheads can be minimized to close to zero.

The \emph{deep pipelining} technique is utilized for FFTs and IFFTs in order to improve throughput and energy efficiency, as illustrated in Fig. \ref{fig:struction}. For example, if a 128-point FFT is implemented as the basic computing block in FPGA, it needs 7 pipeline stages plus 4 additional stages corresponding to memory reading and writing. When IFFT is implemented on such basic computing block, 2 additional stages are needed corresponding to the preprocessing, and biasing and ReLU activation. The element-wise multiplications and additions in the second phase are also pipelined.

One clear advantage of the FPGA-based hardware implementation is the ability of \emph{resource re-use}. Besides the effective time multiplexing of FFTs and IFFTs on the same hardware, the hardware multipliers utilized in the second phase can also re-use those in the FFT computing block. This effective resource re-use can be automatically determined in the FPGA synthesis process (\cite{quartus2}), which could improve the area and energy efficiency of FPGA implementations.

\subsection{Algorithm-Hardware Co-Optimizations}

\begin{figure}
  \centering
  \includegraphics[width=0.24\textwidth]{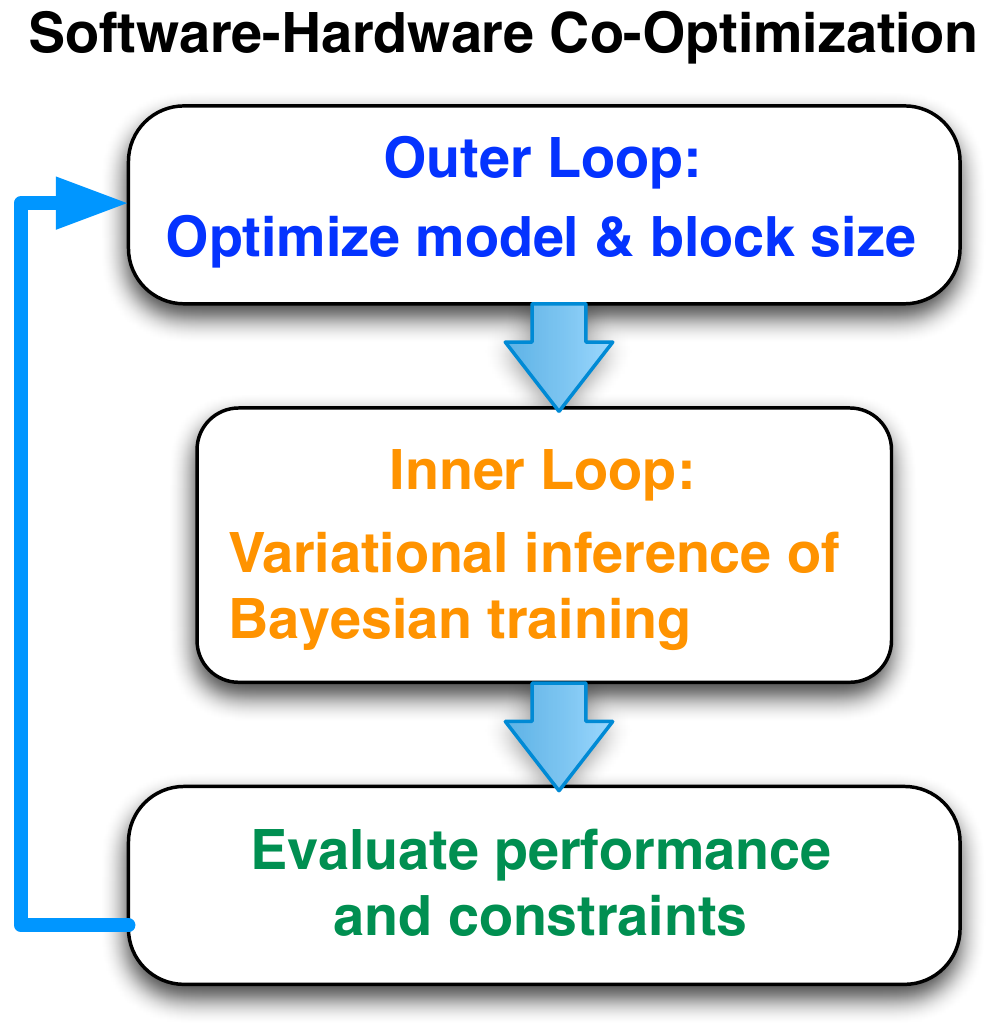}
  \caption{\footnotesize\bf The software-hardware co-optimization of model and block size, and Bayesian training.}
  \label{fig:coop}
\end{figure}

Finally, an algorithm-hardware co-optimization framework is developed, which comprises (i) model selection and optimization, (ii) hardware optimization, and (iii) variational inference-based Bayesian learning. The overall objective is to maximize the performance (throughput) and energy efficiency of FPGA hardware implementation subject to certain accuracy requirements. More specifically, the first aspect determines the proper block size and weight matrix size, in order to facilitate FPGA-based FFT implementations while satisfying the overall accuracy requirement. For state-of-the-art FPGAs, a proper block size ranges from 64 to 256 (should better be a power of 2) for FC layers and may be smaller for CONV layers. The second aspect includes the exploitation of FFTs with real-valued inputs, i.e., the FFT results of a real-valued vector is symmetric except for the base (first) component (\cite{oppenheim1999discrete}). Because both $\mathbf{x}_j$ and $\mathbf{w}_{ij}$ are real-valued vectors, we only need to store the first half of vectors $\text{FFT}(\mathbf{x}_j)$ and $\text{FFT}(\mathbf{w}_{ij})$, which significantly reduce the storage requirement and computations required in element-wise multiplications. The last aspect uses the variational inference process of Bayesian learning (\cite{blei2006variational}), which is compatible with the proposed framework and can result in accuracy and robustness enhancements. 
Bayesian training using variational inference (\cite{blei2006variational}) is an effective training method to enhance accuracy and robustness of machine learning systems, including neural networks. During training phase, it assumes that each weight is a variable that satisfies certain prior distribution at the beginning. For each training sample, it generates a collection of random weights based on the distribution, and learns both the average and variance of each weight variable. The inference phase (implemented in hardware) will be the same, using the average estimate of each weight.
Based on our results, Bayesian training is the most effective for small data training and small-to-medium neural networks.
The algorithm-hardware co-optimization framework is shown in Fig. \ref{fig:coop}. Overall, the proposed FPGA-based implementation can accommodate the whole DNN model using on-chip block memory, thereby significantly improving the overall energy efficiency.

\section{Experimental Results}
In this section, we provide the experimental results on FPGA implementations of the proposed framework on small to medium-scale DNNs, using MNIST, SVHN, and CIFAR-10 benchmarks. Our FPGAs for implementation include the low-power FPGA Intel (Altera) CyClone V 5CEA9, and the one with higher performance Xilinx Kintex-7 XC7K325T. The former one is the default FPGA used in experiments. We compare the performance (throughput), energy efficiency, and accuracy with the best state-of-the-arts including IBM TrueNorth neurosynaptic processor, emerging device (e.g., memristor crossbar) based neuromorphic systems, analog-based neuromorphic systems, and reference FPGA implementations. IBM TrueNorth (\cite{esser2015backpropagation,esser2016convolutional}) is a neuromorphic CMOS chip fabricated in 28nm technology, with 4096 cores each simulating 256 programmable silicon neurons in a time-multiplexed manner. It implements the \emph{spiking neural network}, which is a bio-inspired type of neural networks and benefits from the ability of globally asynchronous implementations. It can accommodate MNIST, SVHN, and CIFAR-10 benchmarks\footnote{Please note that ImageNet is not currently supported by IBM TrueNorth due to the high-degree neural connections.} in the experiments.

First, we provide the comparison results on accuracy, performance (throughput, in kilo-frames per second (kFPS)), and energy efficiency (in kFPS/W) on the three benchmarks, as shown in Table 1. The baselines include IBM TrueNorth processor and reference FPGA implementations of these benchmarks. We provide results of the proposed framework on three DNNs of MNIST data set with different target accuracies, one for SVHN, and two for CIFAR-10 data set. The first two DNNs of the MNIST data set are multi-layer perceptron (MLP) models that achieve 92.9\% and 95.6\% accuracies, respectively. Prior pooling is applied to reduce the input size to 256 and 128, respectively. The third DNN of the MNIST data set is a CNN similar to the LeNet-5 structure (\cite{lecun1995comparison}). The baseline IBM TrueNorth processor also has different implementations with different accuracy levels for the MNIST data set. For the CIFAR-10 data set, the first DNN is a simple CNN structure, whereas the second is a wide ResNet model (\cite{he2016deep}) that can achieve 94.75\% accuracy, only 0.75\% lower than the best state-of-the-art software implementation. We can observe that under the similar accuracy level, the speedup and energy efficiency gain compared with IBM TrueNorth are at least 152X and 71X, respectively. Under the similar accuracy level, the energy efficiency gain is at least 31X compared with the reference FPGA-based implementation that achieves the highest energy efficiency (\cite{umuroglu2016finn}) (using binary neural networks). Besides the reduction in computational complexity, the high suitability of the proposed framework for hardware implementation, and the highly efficient deep pipelined hardware structure, the reasons for such significant gains also include the requirement of increasing neuron numbers for spiking or binary neural networks to achieve the same accuracy as MLP or CNN, and the inherent long latency in spiking neural networks. 

\begin{table*}[t]
\small
  \caption{Comparison results on accuracy, performance, and energy efficiency of the proposed FPGA designs and baselines.}
  \label{table:1}
  \centering
  \begin{tabular}{lllllll}
    \toprule
    Name & Dataset & Platform & Precision & Accuracy & Performance & Energy eff.\\
         &         &          &           &          & (kFPS)      & (kFPS/W) \\
    \midrule
    Proposed MNIST 1 & MNIST & CyClone V & 12 & 92.9\% & $8.6\times 10^4$ & $1.57\times 10^5$\\
    Proposed MNIST 2 & MNIST & CyClone V & 12 & 95.6\% & $2.9\times 10^4$ & $5.2\times 10^4$\\
    Proposed MNIST 3 & MNIST & CyClone V & 12 & 99.0\% & $363$ & 659.5\\
    Proposed SVHN & SVHN & CyClone V & 12 & 96.2\% & $384.9$ & 699.7\\
    Proposed CIFAR-10 1 & CIFAR-10 & CyClone V & 12 & 80.3\% & $1383$ & 2514\\
    Proposed CIFAR-10 2 & CIFAR-10 & CyClone V & 12 & 94.75\% & $13.95$ & 25.4\\
    \midrule
    TrueNorth (\cite{esser2015backpropagation}) & MNIST & TrueNorth & 2 & 99\%+ & 1.0 & 9.26 \\
    TrueNorth (\cite{esser2015backpropagation}) & MNIST & TrueNorth & 2 & 95\% & 1.0 & 250 \\
    TrueNorth (\cite{esser2016convolutional}) & SVHN & TrueNorth & 2 & 96.7\% & 2.53 & 9.85 \\
    TrueNorth (\cite{esser2016convolutional}) & CIFAR-10 & TrueNorth & 2 & 83.4\% & 1.25 & 6.11 \\
    \midrule
    Umuroglu et al. (\cite{umuroglu2016finn}) & MNIST & ZC706 & 1 & 95.8\% & $1.23\times 10^4$ & 1693 \\
    Umuroglu et al. (\cite{umuroglu2016finn}) & SVHN & ZC706 & 1 & 94.9\% & 21.9 & 6.08 \\
    Umuroglu et al. (\cite{umuroglu2016finn}) & CIFAR-10 & ZC706 & 1 & 80.1\% & 21.9 & 6.08 \\
    Alemdar et al. (\cite{alemdar2016ternary}) & MNIST & Kintex-7 & 2 & 98.3\% & 255.1 & 92.59 \\
    \bottomrule
  \end{tabular}
\end{table*}

Next, we provide sample comparison results with emerging device and analog-based implementations. Because the neural networks and applications may be different, we use the equivalent performance in giga-operations per second (GOPS) and energy efficiency in GOPS/W for fair comparisons. The term ``equivalent'' is utilized because we normalize the number of (multiplication and addition) operations to the original matrix-vector multiplication format. The proposed framework achieves around 5.14 Tera OPS/W (TOPS/W) energy efficiency, which outperforms representative latest results using analog computing and emerging devices. For example, (\cite{shafiee2016isaac,song2017pipelayer,lu20151}) achieve 380.7 GOPS/W, 142.9 GOPS/W, and 1.04 TOPS/W, respectively. The reference work can be either manufactured or device modeling based. Performance wise, as analyzed in (\cite{bayat2016sub,liu2016harmonica,li2016heterogeneous}), a matrix-vector multiplication will take around 100ns and it takes around 1$\mu s$ to perform one inference sample on the MNIST data set (with 90\% - 94\% accuracy). Our achieved highest performance (throughput) for the MNIST data set, i.e., 11.6ns per image recognition in CyClone V FPGA or around 4ns per image in Kintex-7 FPGA, is difficult to achieve even using emerging devices and technology. 

\begin{figure}
  \centering
  \includegraphics[width=0.38\textwidth]{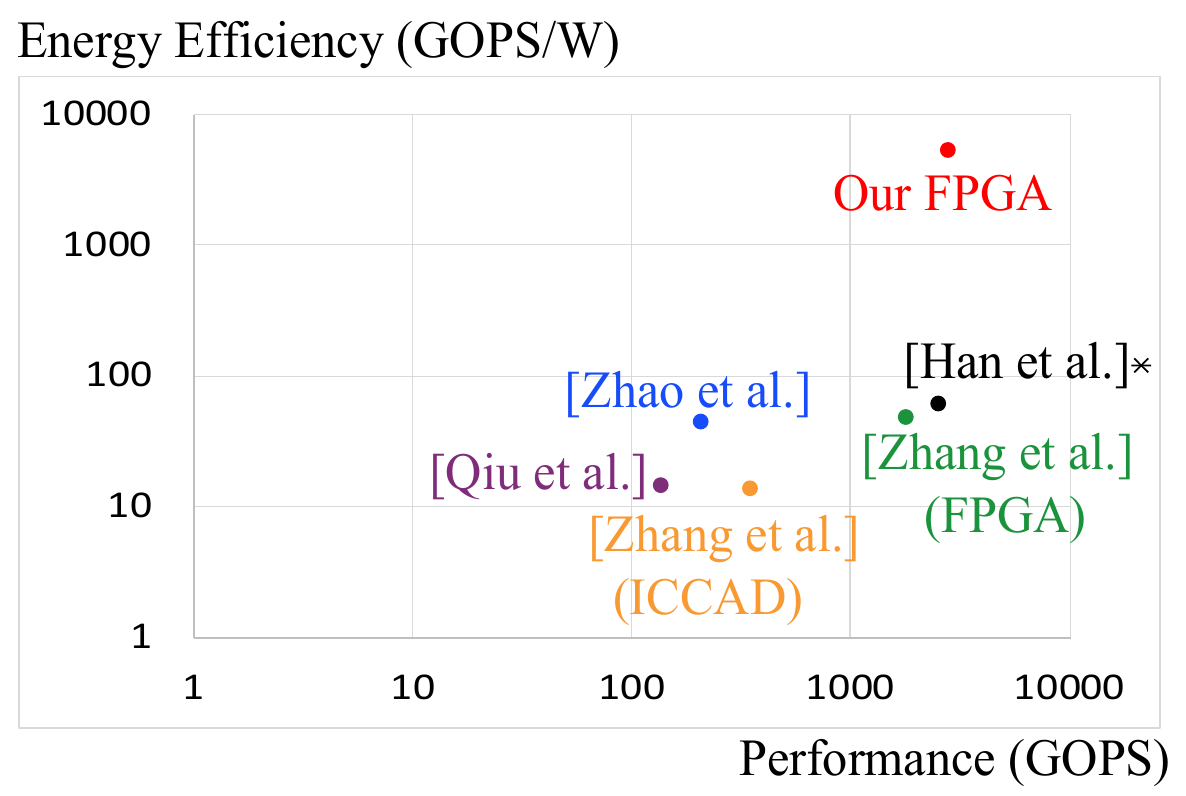}
  \caption{\footnotesize\bf Comparisons of different FPGA implementations on performance (throughput) and energy efficiency.}
  \label{fig:FPGA_NIPS}
\end{figure}

Finally, we provide the comparison results with other FPGA implementations in terms of the equivalent performance (in GOPS) and energy efficiency (in GOPS/W), as shown in Fig. \ref{fig:FPGA_NIPS}. These metrics are relatively fair comparisons although the DNNs for implementations may be different. The baseline FPGA implementations include high-level synthesis-based implementations, implementations of compressed models, etc. A minimum of more than 84X energy efficiency gain can be achieved compared with the reference FPGA implementations. Besides the reduced computational complexity and the high-efficiency hardware implementation, another key reason for such significant energy efficiency gain is because the proposed FPGA-based implementation can accommodate the whole DNN model using on-chip block memory, thereby significantly improving the overall energy efficiency.

\section{Conclusion}
This paper presents an algorithm-hardware co-optimization framework to facilitate ultra high-performance and high energy efficiency hardware implementations of DNNs on FPGAs. The algorithm part adopts the general block-circulant matrices to achieve a fine-grained tradeoff of accuracy and compression ratio. It applies to both FC and CONV layers and contains a mathematically rigorous proof. The proposed algorithm reduces computational complexity per layer from O($n^2$) to O($n\log n$) and storage complexity from O($n^2$) to O($n$), both for training and inference phases. The hardware part consists of highly efficient FPGA-based implementations using effective reconfiguration, batch processing, deep pipelining, resource re-using, and a hierarchical control framework. Experimental results demonstrate that the proposed framework achieves at least 152X speedup in throughput and 71X energy efficiency gain compared with IBM TrueNorth processor under the same test accuracy. It achieves at least 31X energy efficiency gain compared with the reference FPGA-based work.

\section{Acknowledgement}
This work is funded by the National Science Foundation Awards CNS-1650469, CCF-1733701, CNS-1704662, CCF-1657333, CNS-1739748, and CCF-1733834.

\bibliographystyle{aaai}
\bibliography{reference}

\end{document}